\documentclass{article}
\usepackage{arxiv}

\usepackage{geometry}
\usepackage{graphicx}
\usepackage[hidelinks]{hyperref}

\usepackage{amssymb}
\usepackage{bm}
\usepackage{lineno}
\usepackage{amsmath}

\usepackage{multirow}
\usepackage{float}
\usepackage{tabularx}
\usepackage{subcaption}
\usepackage{booktabs}
\usepackage{array,xcolor}
\usepackage{arydshln}

\usepackage{algorithm, algpseudocode}

\usepackage{cite}
\title{Event-driven physics-informed operator learning for reliability analysis}
\author{
  Shailesh Garg  \\
  Department of Applied Mechanics\\
  Indian Institute of Technology Delhi\\
  Hauz Khas, New Delhi 110016, India. \\
  \texttt{shailesh.garg@am.iitd.ac.in} \\
  \And
  Souvik Chakraborty  \\
  Department of Applied Mechanics\\
  Yardi School of Artificial Intelligence (ScAI)\\
  Indian Institute of Technology Delhi\\
  Hauz Khas, New Delhi 110016, India. \\
  \texttt{souvik@am.iitd.ac.in}}
\begin{document}
\maketitle
\begin{abstract}
Reliability analysis of engineering systems under uncertainty poses significant computational challenges, particularly for problems involving high-dimensional stochastic inputs, nonlinear system responses, and multiphysics couplings. Traditional surrogate modeling approaches often incur high energy consumption, which severely limits their scalability and deployability in resource-constrained environments.
We introduce NeuroPOL, \textit{the first neuroscience-inspired physics-informed operator learning framework} for reliability analysis. NeuroPOL incorporates Variable Spiking Neurons into a physics-informed operator architecture, replacing continuous activations with event-driven spiking dynamics. This innovation promotes sparse communication, significantly reduces computational load, and enables an energy-efficient surrogate model. The proposed framework lowers both computational and power demands, supporting real-time reliability assessment and deployment on edge devices and digital twins.
By embedding governing physical laws into operator learning, NeuroPOL builds physics-consistent surrogates capable of accurate uncertainty propagation and efficient failure probability estimation, even for high-dimensional problems. We evaluate NeuroPOL on five canonical benchmarks, the Burgers equation, Nagumo equation, two-dimensional Poisson equation, two-dimensional Darcy equation, and incompressible Navier-Stokes equation with energy coupling. Results show that NeuroPOL achieves reliability measures comparable to standard physics-informed operators, while introducing significant communication sparsity, enabling scalable, distributed, and energy-efficient deployment.
\end{abstract}
\section{Introduction}\label{section: introduction}
Reliability analysis plays a pivotal role in the design and safety assessment of modern engineering systems operating under uncertainty. From aerospace structures and energy infrastructures to autonomous vehicles and smart materials, accurately quantifying the probability of failure is essential for informed decision-making and risk mitigation. However, the increasing complexity of engineering systems, characterized by high-dimensional stochastic inputs, nonlinear responses, and tightly coupled multiphysics phenomena, has rendered conventional reliability analysis approaches computationally prohibitive.
Classical techniques such as the First-Order Reliability Method (FORM) \cite{maier2001first,haldar1995first,luo2024active,wang2022first} and the Second-Order Reliability Method (SORM) \cite{hu2021second,haldar1995first,wang2020confidence} have been extensively studied and successfully applied in many contexts. Yet, accurate estimation of reliability often demands evaluating system responses across a vast space of uncertain inputs, typically using Monte Carlo simulations. While feasible for small-scale systems, this becomes impractical for large-scale, high-fidelity models where each evaluation involves solving complex nonlinear and multiphysics problems, resulting in prohibitive computational costs.
To overcome these challenges, surrogate modeling techniques have been proposed to approximate system responses at a fraction of the computational expense. Popular approaches such as polynomial chaos expansion \cite{blatman2011adaptive,crestaux2009polynomial,zhang2021efficient}, Gaussian process regression (Kriging) \cite{kaymaz2005application,wilkie2021gaussian,wang2025novel}, and the response surface method \cite{roussouly2013new,li2016response} have demonstrated considerable success for low-dimensional, smooth response surfaces. However, these methods suffer from the well-known curse of dimensionality and fail to capture the intricate behaviors of high-dimensional, highly nonlinear, and tightly coupled systems, significantly limiting their applicability in modern reliability analysis.

In recent years, deep learning-based frameworks \cite{rusk2016deep,goodfellow2016deep,lecun2015deep} have emerged as a powerful alternative to traditional surrogate modeling techniques, primarily due to their strong generalization capabilities. 
Broadly, these frameworks can be classified into two categories: (i) data-driven frameworks and (ii) physics-informed frameworks. Within the first category, operator learning algorithms such as the Deep Operator Network (DeepONet) \cite{lu2021learning}, Fourier Neural Operator (FNO) \cite{lifourier}, Wavelet Neural Operator (WNO) \cite{tripura2023wavelet}, and their recent extensions \cite{seidman2022nomad,li2022transformer,tran2021factorized,li2023fourier,guibas2021adaptive,rani2024generative,lei2025u} have demonstrated remarkable performance, particularly in scenarios where abundant, high-quality training data is available but the governing physics is partially known or uncertain. 
In contrast, the second category includes frameworks such as Physics-Informed Neural Networks (PINNs) \cite{raissi2019physics} and their variants \cite{kharazmi2021hp,jagtap2020extended,navaneeth2023stochastic,pang2019fpinns}, which embed physical laws directly into the learning process by incorporating Partial Differential Equations (PDEs) as soft constraints in the loss function. These physics-informed models are inherently more robust and data-efficient, making them particularly suitable for reliability analysis problems where system dynamics are well-characterized by known PDEs, but high-fidelity data is scarce or expensive to generate.
Building on the success of PINNs, several recent studies have extended these ideas to operator learning frameworks trained using the underlying physics. Notable examples include the Physics-Informed Neural Operator (PINO) \cite{li2024physics}, Variational Physics-Informed Operator \cite{eshaghi2025variational}, and Physics-Informed Wavelet Neural Operator (PIWNO) \cite{navaneeth2024physics}. Unlike vanilla PINNs, which are trained for a single set of input conditions, these physics-informed operator networks learn mappings between functional inputs and outputs, enabling efficient generalization across varying input scenarios.
However, despite these advantages, all existing physics-informed operator frameworks suffer from a critical bottleneck: they are built on conventional deep learning architectures with continuous activation functions and dense, high-precision computations, which make them extremely energy-intensive during inference. This limitation becomes especially severe in real-world deployment scenarios where computational resources are scarce—for example, edge devices, embedded systems, and onboard computing platforms in aerospace, autonomous vehicles, or smart infrastructure monitoring. In such environments, the high power demand of conventional physics-informed neural operators directly hampers their deployability, making them impractical for scenarios that demand real-time, low-power decision-making.

To overcome this fundamental limitation, we introduce the first-ever Neuroscience-inspired, energy-efficient Physics-informed Operator Learning framework (NeuroPOL) that leverages A Variable Spiking Neuron \cite{garg2024neuroscience,garg2023neuroscience} (VSN) model to enable lower power consumption without compromising predictive performance. Unlike conventional deep learning architectures, which rely on continuous activation functions and dense floating-point computations, our framework harnesses the event-driven dynamics of spiking neurons, where computations are triggered only when spikes occur. This results in highly sparse, low-latency inference and makes our framework inherently suitable for real-time deployment on embedded and edge devices operating under severe energy constraints.
Among various neuroscience-inspired spiking neuron models like the leaky integrate and fire model \cite{yamazaki2022spiking,tavanaei2019deep}, the choice of VSNs is supported by their performance in regression tasks while promoting sparse communication. VSNs communicate through non-binary spikes and, through their dynamics, take advantage of both intermittent firing and continuous activation. This design achieves higher accuracy than LIF neurons in regression tasks \cite{garg2024neuroscience,jain2025hybrid,garg2024distribution,garg2023neuroscience}, while remaining more energy-efficient than conventional continuous activations. 
What truly sets this work apart is its ability to seamlessly integrate the rigor of physics-informed learning with the efficiency of spiking neural computation. By embedding PDE constraints directly into a spiking-based operator learning architecture, our framework achieves the robustness, generalization capability, and data efficiency of physics-informed operator methods while radically improving energy efficiency and deployability. This breakthrough represents a paradigm shift in scientific machine learning, enabling the development of next-generation digital twins capable of autonomous, real-time operation in harsh, resource-limited environments such as edge sensors for smart infrastructures, embedded systems for aerospace platforms, and IoT-enabled monitoring networks.

We rigorously evaluate the proposed {NeuroPOL} framework across five canonical mechanics problems, the \textit{Burgers equation}, the \textit{Nagumo equation}, a \textit{two-dimensional Poisson equation},  a \textit{two-dimensional Darcy equation}, and a \textit{two-dimensional incompressible Navier-Stokes equation coupled with the energy equation}, to establish its accuracy, efficiency, and deployability. The predictive performance of NeuroPOL is benchmarked against the \textit{vanilla Physics-Informed Wavelet Neural Operator} and validated against \textit{high-fidelity Monte Carlo simulations}, which is widely regarded as the {gold standard} for reliability analysis. Across all test cases, NeuroPOL consistently matches the predictive accuracy and reliability measures of its vanilla counterparts while introducing significant communication sparsity and achieving {improvements in energy efficiency}. The salient contributions of this work can be summarized as follows:
\begin{itemize}
    \item \textbf{First neuroscience-inspired physics-informed operator learning framework}: We propose \textbf{NeuroPOL}, the first neuroscience-inspired and physics-informed neural operator. Leveraging the event-driven nature of spiking architectures, NeuroPOL achieves \textbf{sparse and energy-efficient computation} while preserving state-of-the-art predictive accuracy. This unique combination makes NeuroPOL inherently suitable for \textbf{resource-constrained and edge-deployable scientific computing}.
    
    \item \textbf{Physics-constrained generalization}: The framework \textbf{seamlessly embeds PDE-based physics constraints} within a neuroscience-inspired operator, ensuring \textbf{robust generalization across varying input conditions} and improving scalability compared to traditional data-driven models.
    \item \textbf{Stochastic projection for gradient computation in loss evaluation}:  
    To accommodate the discontinuous nature of spiking activations within a physics informed framework, NeuroPOL employs a stochastic projection method for gradient estimation during loss function computation. This allows the model to compute physics-informed gradients reliably, even in the presence of non-smooth neuron dynamics.
\end{itemize}

The rest of the paper is arranged as follows: Section \ref{section: problem statement} discusses the problem statement, and Section \ref{section: proposed framework} discusses the proposed NeuroPOL surrogate. Section \ref{section: numerical} discusses the various examples, and Section \ref{section: conclusion} concludes the findings of the manuscript.
\section{Problem Statement}\label{section: problem statement}
Let $(\Xi,\mathcal{F},\mathbb{P})$ be a probability space and let 
$\boldsymbol{\xi}:\Xi\!\to\!\mathbb{R}^d$ denote a vector of uncertain inputs with law 
$\mu := \mathbb{P}\!\circ\!\boldsymbol{\xi}^{-1}$. 
Let $\Omega\subset\mathbb{R}^n$ be a bounded Lipschitz domain with boundary $\partial\Omega$.
For each realization $\boldsymbol{\xi}\in\mathbb{R}^d$, the system state 
$u(\cdot;\boldsymbol{\xi})$ is the (weak) solution in a Hilbert space 
$\mathcal{V}\subset H^1(\Omega)^k$ of the parameterized boundary value problem
\begin{equation}
\label{eq:pde}
\begin{cases}
\mathcal{L}\!\left(u(\bm{x},t;\boldsymbol{\xi});\,\bm{x},t,\boldsymbol{\xi}\right)=
f(\bm{x},t;\boldsymbol{\xi}) & \text{in }\Omega \text{ at } t\in (0,T],\\[2pt]
\mathcal{L}_b\!\left(u(\bm{x},t;\boldsymbol{\xi});\,\bm{x},t,\boldsymbol{\xi}\right)=
g_b(\bm{x},t;\boldsymbol{\xi}) & \text{on }\partial\Omega \text{ at } t\in (0,T],\\[2pt]
\mathcal{L}_t\!\left(u(\bm{x};\boldsymbol{\xi});\,\bm{x},\boldsymbol{\xi}\right)=
g_t(\bm{x};\boldsymbol{\xi}) & \text{in }\Omega \text{ at } t=0,
\end{cases}
\end{equation}
where $\mathcal{L}$ may be nonlinear and encode multiphysics couplings, $\mathcal{L}_b$ collects boundary/interface conditions and $\mathcal{L}_t$ collects initial conditions. 
We write the (deterministic) solution operator as
\begin{equation}
\label{eq:solution-operator}
\mathcal{T}:\;\mathcal{X}\to\mathcal{V}, 
\qquad 
u(\cdot;\boldsymbol{\xi})=\mathcal{T}\!\big(a(\bm x ,t;\boldsymbol{\xi})\big),
\end{equation}
with input $a(\bm x,t ;\boldsymbol{\xi})\in\mathcal{X}$ aggregating coefficients, sources, and boundary data.
For time-independent relaibility, let $\mathcal{Q}:\mathcal{V}\to\mathbb{R}^p$ be a (possibly nonlinear) quantity-of-interest (QoI) operator and 
$G:\mathbb{R}^p\to\mathbb{R}$ a limit-state map. 
Define the performance margin
\begin{equation}
m(\boldsymbol{\xi}) \;=\; G\!\left(\mathcal{Q}\!\big(\mathcal{T}(a(\bm x ;\boldsymbol{\xi}))\big)\right),
\end{equation}
the failure set $\mathcal{F}:=\{\boldsymbol{\xi}\in\mathbb{R}^d: m(\boldsymbol{\xi})\le 0\}$,
and the failure probability
\begin{equation}
\label{eq:pf}
p_f \;=\; \mathbb{P}\!\left(m(\boldsymbol{\xi})\le 0\right)
\;=\; \int_{\mathbb{R}^d} \mathbf{1}\!\left\{ G\!\left(\mathcal{Q}\!\big(\mathcal{T}(a(\bm x ;\boldsymbol{\xi}))\big)\right) \le 0 \right\} \,\mathrm{d}\mu(\boldsymbol{\xi}),
\end{equation}
where $\mathbf{1}\left\{\cdot\right\}$ represents the indicator function.
For i.i.d.\ samples $\{\boldsymbol{\xi}_i\}_{i=1}^N\sim\mu$, the gold-standard estimator is
\begin{equation}
\widehat{p}_f^{\,\mathrm{MC}} \;=\; \frac{1}{N}\sum_{i=1}^{N}
\mathbf{1}\!\left\{ G\!\left(\mathcal{Q}\!\big(\mathcal{T}(a(\boldsymbol{\xi}_i))\big)\right)\le 0 \right\}.
\end{equation}
For time-dependent reliability, we are interested in the random variable $\tau(\boldsymbol{\xi})$, denoting the \emph{first time-to-failure}
\begin{equation}
    \tau(\boldsymbol{\xi}) = \inf \Big\{ t \in [0,T] \,:\, m(\boldsymbol{\xi}) \leq 0 \Big\}.
\end{equation}
The survival function $S(t)$, i.e., the probability that the system survives beyond time $t$, is given by:
\begin{equation}\label{eq:st}
    S(t) = \mathbb{P}\big[\tau(\boldsymbol{\xi}) > t\big] 
    = \int_{\{\boldsymbol{\xi}: \, G(u(\boldsymbol{x},t';\boldsymbol{\xi}),\boldsymbol{\xi}) > 0, \ \forall \ t' \in [0,T]\}} \rho_{\boldsymbol{\xi}}(\boldsymbol{\xi}) \, d\boldsymbol{\xi}.
\end{equation}
Similar to Eq. \eqref{eq:pf}, Eq. \eqref{eq:st} can also be approximated using MCS.
To compute $S(t)$, we define the \emph{survival indicator} for sample $i$ as
\begin{equation}
\label{eq:survival-indicator}
I_i^{(\Delta t)}(t) 
\;:=\; \prod_{k=0}^{K} \mathbf{1}\!\left\{\, G\!\left(\mathcal{Q}\!\big(\mathcal{T}(a(\bm x,t ;\boldsymbol{\xi}_i))\big)\right) > 0 \,\right\}
\;\in\;\{0,1\},
\end{equation}
which equals $1$ iff the trajectory remains in the safe domain at all grid points up to time $t$. 
A MC estimator of the survival function $S(t)$ is then
\begin{equation}
\label{eq:MC-survival}
\widehat{S}_{N}^{(\Delta t)}(t)
\;=\; \frac{1}{N}\sum_{i=1}^{N} I_i^{(\Delta t)}(t).
\end{equation}
The corresponding estimator of the time-dependent failure probability is
\begin{equation}
\label{eq:MC-failure}
\widehat{P}_{f,N}^{(\Delta t)}(t) \;=\; 1 - \widehat{S}_{N}^{(\Delta t)}(t).
\end{equation}
In this work, we are interested in computing the probability density function of the first time-to-failure for time-dependent problems and the probability of failure for time-independent problems.
In both cases, each indicator evaluation requires solving Eq.~\eqref{eq:pde}, which is prohibitive for large-scale, nonlinear, or multiphysics systems and large $N$. Therefore,
we seek a parametric operator surrogate $\mathcal{S}_\theta:\mathcal{X}\to\mathcal{V}$ such that
\begin{subequations}
\begin{equation} 
\label{eq:surrogate-estimator}
\widehat{p}_f(\theta) \;=\; \frac{1}{N}\sum_{i=1}^{N}
\mathbf{1}\!\left\{ G\!\left(\mathcal{Q}\!\big(\mathcal{S}_\theta(a(\bm x ;\boldsymbol{\xi}_i))\big)\right)\le 0 \right\},
\end{equation}
\begin{equation}
\label{eq:survival-indicator_sur}
I_i^{(\Delta t)}(t) 
\;:=\; \prod_{k=0}^{K} \mathbf{1}\!\left\{\, G\!\left(\mathcal{Q}\!\big(\mathcal{S}_\theta(a(\bm x,t ;\boldsymbol{\xi}_i))\big)\right) > 0 \,\right\}
\;\in\;\{0,1\},
\end{equation}
\end{subequations}
provides an accurate and \emph{deployable} estimate of Eq.~\eqref{eq:pf} at low latency and low energy. Additionally, it is desirable that the number of calls to Eq. \eqref{eq:pde} be minimized in order to ensure a tractable solution.
\section{Neuroscience-inspired Physics-informed Operator Learning}\label{section: proposed framework}
In this section, we propose \emph{NeuroPOL}, a neuroscience-inspired and physics-informed neural operator framework for efficient and accurate estimation of reliability metrics such as probability of failure $p_f$ and PDF of the first time-to-failure $\tau(\boldsymbol{\xi})$.  
The key idea is to replace the high-fidelity solution operator $\mathcal{T}$ in Eq.~\eqref{eq:solution-operator} with a parametric neural operator surrogate $\mathcal{S}_\theta:\mathcal{X}\to\mathcal{V}$, trained to approximate the mapping $a(\bm x,t;\boldsymbol{\xi}) \mapsto u(\cdot;\boldsymbol{\xi})$ at significantly reduced computational cost.
In the proposed NeuroPOL framework, we employ the \emph{Wavelet Neural Operator} (WNO) as the base architecture for $\mathcal{S}_\theta$, although other neural operator variants, such as the Fourier Neural Operator (FNO), can also be substituted without loss of generality.  
Neural operators form a class of learning architectures that approximate mappings between infinite-dimensional function spaces.  
Formally, given an input function $a(\boldsymbol{x,t};\boldsymbol{\xi})\in\mathcal{X}$ defined on $\boldsymbol{x}\in\Omega\subset\mathbb{R}^n$, the surrogate solution $u_\theta(\cdot;\boldsymbol{\xi})=\mathcal{S}_\theta(a(\bm x,t;\boldsymbol{\xi}))\in\mathcal{V}$ is obtained via the WNO architecture as:
\begin{equation}
\begin{gathered}
    L_0 = \mathcal{U}\big(a(\boldsymbol{x},t;\boldsymbol{\xi})\big), 
    \quad L_0\in\mathbb{R}^{d_u}, \\[6pt]
    L_i = \sigma\!\Big( \mathcal{W}^{-1}\!\big( R \, \mathcal{W}(L_{i-1}) \big)(\boldsymbol{x}) + W L_{i-1} \Big), 
    \quad i=1,\dots,n, 
    \quad L_i\in\mathbb{R}^{d_u}, \\[6pt]
    u_\theta(\cdot;\boldsymbol{\xi}) = \mathcal{P}(L_n),
\end{gathered}
\label{eq:wno_architecture}
\end{equation}
where $\sigma(\cdot)$ is a nonlinear activation, $\mathcal{U}:\mathcal{X}\to\mathbb{R}^{d_u}$ and $\mathcal{P}:\mathbb{R}^{d_u}\to\mathcal{V}$ denote the \emph{uplifting} and \emph{projection} operators, respectively, both modeled using fully-connected neural networks.  
The operator $W:\mathbb{R}^{d_u}\to\mathbb{R}^{d_u}$ represents a learnable linear transformation, while $R$ is a parameterized kernel responsible for nonlocal mixing in the latent space.  

A distinguishing feature of WNO is its use of the continuous wavelet transform to capture multiscale, spatially localized representations of the latent field.  
For a function $f\in\mathbb{R}^{d_u}$, the forward and inverse wavelet transforms, $\mathcal{W}(\cdot)$ and $\mathcal{W}^{-1}(\cdot)$, are defined as:
\begin{equation}
\begin{gathered}
\mathcal{W}(f)(s,t_r) 
= f_w(s,t_r) 
= \int_{\Omega} f(x) \,\dfrac{1}{|s|^{1/2}}\,\psi\!\left(\dfrac{x-t_r}{s}\right)\,\mathrm{d}x, \\[8pt]
\mathcal{W}^{-1}(f_w)(x) 
= \dfrac{1}{C_\psi}\int_0^\infty\!\int_{\Omega} 
f_w(s,t_r)\,\dfrac{1}{|s|^{1/2}}\,\widetilde{\psi}\!\left(\dfrac{x-t_r}{s}\right)
\,\mathrm{d}t_r\,\dfrac{\mathrm{d}s}{s^2},
\end{gathered}
\label{eq:wavelet_transform}
\end{equation}
where $\psi\in\mathcal{L}^2(\mathbb{R})$ is the orthonormal \emph{mother wavelet}, localized in both spatial and frequency domains, and $\widetilde{\psi}$ is its dual.  
Here $s\in\mathbb{R}^+$ is the scale parameter, $t_r\in\mathbb{R}$ is the translation parameter, and $C_\psi\in\mathbb{R}^+$ is the \emph{admissibility constant}, given by:
\begin{equation}
    C_\psi = 2\pi\int_{\Omega}\dfrac{|\psi_\omega|^2}{|\omega|}\,\mathrm{d}\omega,
\end{equation}
where $\psi_\omega$ is the Fourier transform of $\psi$.  
By exploiting the localized multiresolution properties of wavelet bases, the WNO efficiently captures both fine-scale and global features of the solution operator. 

In order to achieve sustainable and energy-efficient operator learning, we incorporate Variable Spiking Neurons (VSNs) into the proposed NeuroPOL framework, replacing the conventional continuous activations $\sigma(\cdot)$ used in standard dense architectures. Unlike continuous activations, which fire deterministically for every incoming input and consequently require executing all subsequent synaptic operations, VSNs introduce an event-driven computation paradigm where synaptic operations are triggered only when membrane potentials cross a learnable firing threshold.
Formally, let $L_i$ denote the $i$-th layer of the network and $\bm z_{\bar t} \in \mathbb{R}^{d}$ be the presynaptic input current received at spike time step $\bar t$. The dynamics of the VSNs are defined as
\begin{equation}\label{eq:vsn_dyn}
\begin{gathered}
\bm M_{\bar t} = \beta \bm M_{\bar t-1} + \bm z_{\bar t}, \\
\bm y_{\bar t} =
\begin{cases}
\bm 0, & \text{if } \bm M_{\bar t} < \mathcal{T}_h, \\
\sigma\left(\bm z_{\bar t}\right), & \text{if } \bm M_{\bar t} \geq \mathcal{T}_h,
\end{cases}
\end{gathered}
\end{equation}
where $\bm M_{\bar t} \in \mathbb{R}^{d}$ denotes the membrane potential at spike time $\bar t$, $\bm y_{\bar t} \in \mathbb{R}^{d}$ is the output spike, and $\beta \in (0,1)$ is the membrane leakage coefficient. The threshold $\mathcal{T}_h \in \mathbb{R}^d$ controls the firing activity of the neuron and can either be fixed as a hyperparameter or learned jointly with network parameters.

The event-driven nature of VSNs leads to a sparse activation pattern across time steps, effectively reducing the number of required Multiply-Accumulate (MAC) operations compared to a dense, continuously firing counterpart. Let $\text{MAC}_{\theta}^{\mathrm{dense}}(\cdot)$ and $\text{MAC}_{\theta}^{\mathrm{vsn}}(\cdot)$ denote the number of MAC operations for the dense and spiking variants, respectively, for a given input $a (\bm x; \bm \xi)$. Then, by construction,
\begin{equation}
\text{MAC}_{\theta}^{\mathrm{vsn}}(\cdot) \ll \text{MAC}_{\theta}^{\mathrm{dense}}(\cdot),
\end{equation}
resulting in significant computational and energy savings while maintaining the expressivity of the underlying neural operator. Since the sparse communication results in reduced MAC operations when using VSNs, in this work, we report neuron activity or spiking activity of VSNs as a surrogate measure for energy efficiency. When the spiking rate is below 100\%, the NeuroPOL will prove to be energy efficient when deployed with neuromorphic hardware.

To reduce the data-dependency of the neuroscience-inspired neural operator, we propose to train the underlying model using physics-informed loss function by exploiting the governing physics in Eq.~\eqref{eq:pde}. In essence, we approximate  the exact solution operator 
$\mathcal{T}:\mathcal{X}\to\mathcal{V}$ in Eq.~\eqref{eq:solution-operator} 
with a parametric surrogate operator $\mathcal{S}_\theta:\mathcal{X}\to\mathcal{V}$, 
\begin{equation}
    \hat{u}(\cdot;\boldsymbol{\xi}) := \mathcal{S}_\theta\!\big(a(\bm x,t ; \boldsymbol{\xi}),\bm x\big)\in\mathcal{V},
\end{equation}
where $\theta\in\Theta\subset\mathbb{R}^P$ denotes the learnable parameters of the surrogate model.
We note that $\mathcal{S}_\theta$  is implemented using WNO combined with VSNs discussed above. 
To ensure that $\mathcal{S}_\theta$ respects the underlying PDE structure, we minimize a composite loss function
\begin{equation}
\label{eq:pino-loss}
\mathcal{L}(\theta)
    = \lambda_\mathrm{data}\,\mathcal{L}_\mathrm{data}(\theta)
    + \lambda_\mathrm{PDE}\,\mathcal{L}_\mathrm{PDE}(\theta)
    + \lambda_\mathrm{BC}\,\mathcal{L}_\mathrm{BC}(\theta)
    + \lambda_\mathrm{IC}\,\mathcal{L}_\mathrm{IC}(\theta),
\end{equation}
where $\lambda_\mathrm{data}$, $\lambda_\mathrm{PDE}$, $\lambda_\mathrm{BC}$, and $\lambda_\mathrm{IC}$ 
are positive weights controlling the relative contribution of each term:
\begin{align}
\mathcal{L}_\mathrm{data}(\theta)
    &= \frac{1}{N_u}\sum_{i=1}^{N_u} 
    \left\| \mathcal{S}_\theta\!\big(a(\boldsymbol{\bm x}, t ; \bm{\xi}_i), \bm x, t\big) - u(\cdot;\boldsymbol{\xi}_i) \right\|_{{2}}^2,\\[4pt]
\mathcal{L}_\mathrm{PDE}(\theta)
    &= \frac{1}{N_r}\sum_{j=1}^{N_r} 
    \left\| \mathcal{L}\!\Big(\mathcal{S}_\theta(a(\bm x, t; \boldsymbol{\xi}_j),\bm{x}, t);\bm{x}, t,\boldsymbol{\xi}_j\Big)
    - f(\bm{x},t;\boldsymbol{\xi}_j) \right\|_2^2,\\[4pt]
\mathcal{L}_\mathrm{BC}(\theta)
    &= \frac{1}{N_b}\sum_{k=1}^{N_b} 
    \left\| \mathcal{B}\!\Big(\mathcal{S}_\theta(a(\bm x,t; \boldsymbol{\xi}_k), \bm{x},t);\bm{x},t,\boldsymbol{\xi}_k\Big)
    - g_b(\bm{x},t;\boldsymbol{\xi}_k) \right\|_2^2,\\[4pt]
\mathcal{L}_\mathrm{IC}(\theta)
    &= \frac{1}{N_b}\sum_{l=1}^{N_b} 
    \left\| \mathcal{B}\!\Big(\mathcal{S}_\theta(a(\bm x,0; \boldsymbol{\xi}_k), \bm{x},0);\bm{x},0,\boldsymbol{\xi}_l\Big)
    - g_t(\bm{x};\boldsymbol{\xi}_l) \right\|_2^2.
\end{align}
Here, $\mathcal{L}_\mathrm{data}$ enforces agreement with available solution snapshots,
$\mathcal{L}_\mathrm{PDE}$ enforces the PDE residual from Eq.~\eqref{eq:pde}, $\mathcal{L}_\mathrm{BC}$ enforces the boundary/interface conditions,
and $\mathcal{L}_\mathrm{IC}$ enforces the initial conditions.
The overall framework combining WNO, VSN, and physics-informed loss function is referred to as the Neuroscience-inspired Physics-informed Operator Learning (NeuroPOL). 
Once trained, $\mathcal{S}_\theta$ serves as a fast surrogate for evaluating 
the QoIs $\mathcal{Q}(\mathcal{S}_\theta(a(\bm x,t;\boldsymbol{\xi})))$ needed for computing the failure probability in Eq.~\eqref{eq:pf} and the time-dependent survival function in Eq.~\eqref{eq:st}.  
Overall, NeurPOL drastically reduce 
the number of full PDE solves required for training and ensures energy efficiency, while achieving accurate and physically consistent 
surrogates for large-scale reliability computations.

To compute the loss function associated with NeuroPOL, it is necessary to evaluate the gradients of the network output with respect to the inputs. However, the introduction of VSNs induces discontinuities in the network’s activation dynamics, making the use of standard backpropagation infeasible for this purpose. To address this, we adopt the \emph{stochastic projection method} to approximate the required spatial and temporal derivatives of the NeuroPOL outputs.  
Specifically, given a query location $\bar{\bm x} = (\bm x_k,t)$ in the spatio-temporal domain $\Omega \times \mathcal{T}$, we define a local neighborhood of radius $r_n$ around $\bar{\bm x}$ and randomly sample $N_b$ collocation points $\{\bm x_i\}_{i=1}^{N_b}$ within this region. The gradient of the predicted solution with respect to $\bar{\bm x}$ is then approximated as:
\begin{equation}
\frac{\partial \mathcal{S}_\theta(a(\bm x,t ;\boldsymbol{\xi}_k), \bm {\bar x})}{\partial \bar{\bm x}} 
= 
\frac{
\displaystyle \frac{1}{N_b} \sum_{i=1}^{N_b} 
\Big(\mathcal{S}_\theta(a(\bm x,t ;\boldsymbol{\xi}_k), \bm x_i,t) - \mathcal{S}_\theta(a(\bm x,t ;\boldsymbol{\xi}_k), \bar{\bm x}) \Big)
(\bm x_i - \bar{\bm x})^\top
}
{
\displaystyle \frac{1}{N_b} \sum_{i=1}^{N_b} 
(\bm x_i - \bar{\bm x})(\bm x_i - \bar{\bm x})^\top
}.
\label{eq:sp_method_grad}
\end{equation}
It is important to emphasize that the stochastic projection method is employed \emph{exclusively} for computing the gradients of the network outputs required in the physics-informed loss function. For computing the gradients of the loss function with respect to the trainable parameters $\bm \theta_W$, standard backpropagation is still used. However, due to the non-differentiability of the thresholding operation in the VSN dynamics, we leverage \emph{surrogate gradients} to enable effective training. In particular, we replace the derivative of the discontinuous threshold function 
with that of a smooth surrogate function. In this work, we adopt the \emph{fast sigmoid} as the surrogate activation, whose derivative is substituted in place of $\frac{\partial \mathcal{H}(u - \mathcal{T})}{\partial u}$ during backpropagation, where $\mathcal{H}(\cdot)$ denotes the Heaviside step function implicit in the spiking mechanism of VSNs.
\begin{algorithm}[ht!]
\caption{Training algorithm for NeuroPOL}
\label{algo}
\vspace{0.35em}
\textbf{Requirements:} PDE operators $\mathcal{N}$, $\mathcal{N}_b$, and $\mathcal{N}_t$. PDE domain $\Omega$ and boundary $\partial \omega$, time interval $[0,T]$, and full spatio‐temporal grid $\{(\bm x_i,t_k)\}\subset D\times[0,T]$.
\begin{algorithmic}[1]
\State \textbf{Initialize:} 
Trainable parameters $\bm \theta_W$ of NeuroPOL, including threshold and leakage parameters of VSN layers, if considered trainable. 
\For{number of epochs}  
  \State Compute NeuroPOL output corresponding to full spatio temporal grid and the given boundary and initial conditions.   
  \State Use Eq. \eqref{eq:sp_method_grad} to compute gradients for the loss function defined in Eq. \eqref{eq:pino-loss}.
  \State Use backpropagation with surrogate gradients for VSN layers to compute the gradients of loss function with respect to the trainable parameters of NeuroPOL.  
  \State Update $\bm \theta_W$ using gradients computed in previous step and suitable optimizer algorithm. 
\EndFor
\State \textbf{Output:} Trained VS‑PIWNO model, that can be used to generalize for new inputs.
\end{algorithmic}
\end{algorithm}
Training process for the NeuroPOL architecture is given in Algorithm \ref{algo}.
\begin{figure}[ht!]
    \centering
    \includegraphics[width=\linewidth]{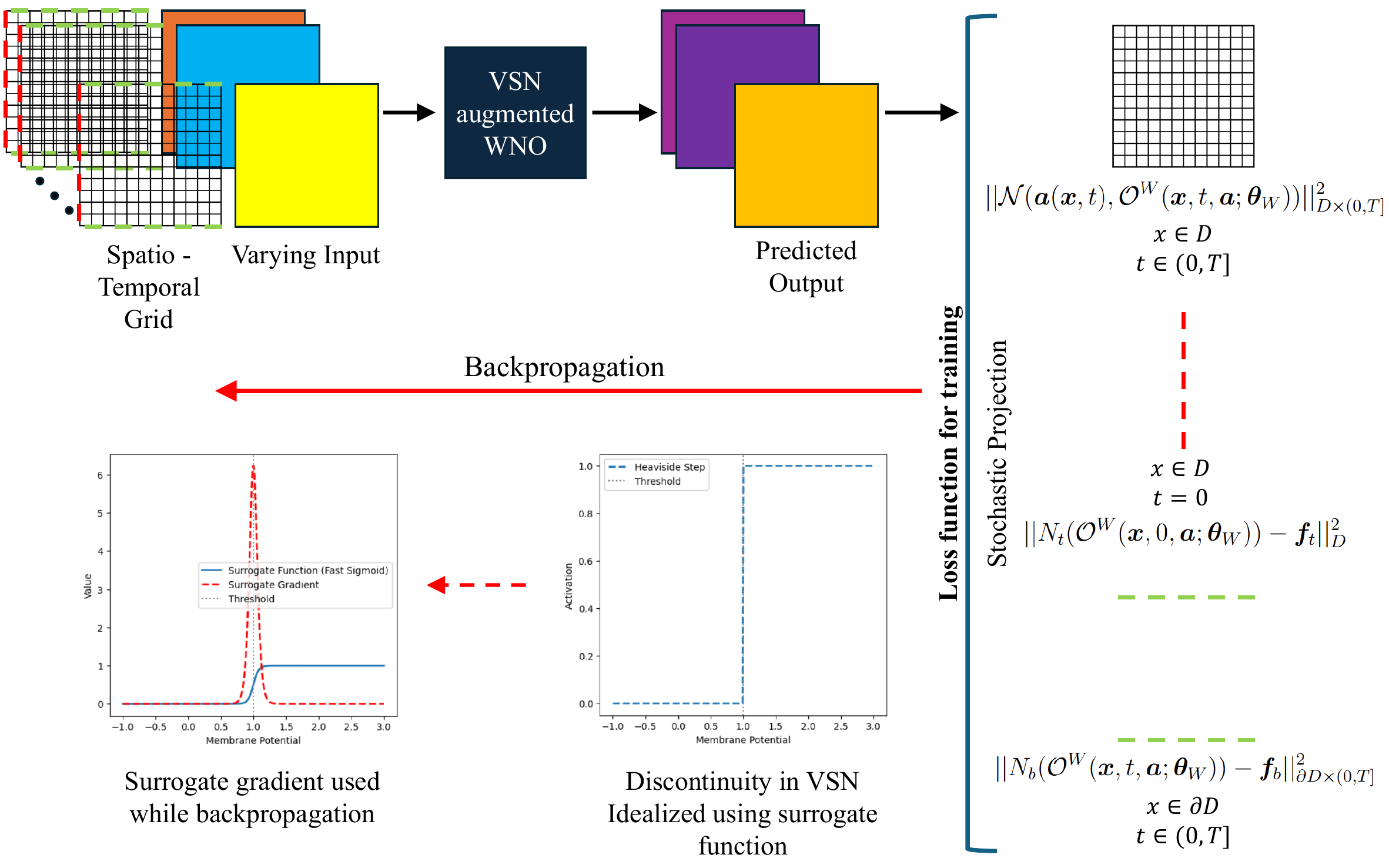}
    \caption{Schematic for the information flow during NeuroPOL training. The model takes spatio-temporal inputs defined on a discretized grid and produces outputs using the VSN augmented WNO architecture. Training involves multiple loss components, including data loss and physics based loss. To enable gradient-based optimization, discontinuities in the VSN are approximated using a surrogate functions, and their gradients are used during backpropagation.}
    \label{fig: vspiwno flow}
\end{figure}
A schematic for the information flow in NeuroPOL is given in Fig. \ref{fig: vspiwno flow}.
\begin{algorithm}[ht!]
\caption{NeuroPOL as surrogate for reliability analysis}
\label{alg:pio_reliability}
\textbf{Requirements:} Trained NeuroPOL framework obtained using Algorithm 1, limit-states for the given problem, and samples of the input conditions.  
\begin{algorithmic}[1]
\For{$i = 1, \ldots, N$}
    \State Draw $i$-th sample from the distribution of the input functions.
    \State Using the trained NeuroPOL framework, obtain predictions.
    \State Evaluate the limit state function.
\EndFor
\State Obtain the probability of failure based on the makeup of the problem, viz, time-dependent or time-independent.
\end{algorithmic}
\textbf{Output:} Probability density function, Failure probability and Reliability Index.
\end{algorithm}
Algorithm \ref{alg:pio_reliability} lays out the steps for reliability analysis using NeuroPOL as surrogate framework.
Once trained, $\mathcal{S}_\theta$ enables low-latency, energy-efficient evaluation of reliability metrics such as $\widehat{p}_f(\theta)$ in Eq.~\eqref{eq:surrogate-estimator} and survival probabilities $\widehat{S}_N^{(\Delta t)}(t)$ in Eq.~\eqref{eq:MC-survival}, facilitating efficient computation of the full probability density function of $\tau(\boldsymbol{\xi})$.

\subsection{Input Encoding}
Since we deploy VSNs within NeuroPOL, the incoming information can be in the form of spikes occuring at different spike time steps $\bar t$. This can be collected at source in this format using sensors like event-driven cameras, or continuous values can be encoded using techniques like rate encoding or time encoding \cite{kim2022rate,auge2021survey}. However, another approach in spiking neural networks is using the inputs directly without encoding. This retains the maximum information and authors in their previous works \cite{garg2023neuroscience,garg2024neuroscience,jain2025hybrid,garg2024distribution} observed that this approach leads to the best accuracy in regression tasks. 

\section{Numerical Illustrations}\label{section: numerical}
To assess the effectiveness of the proposed \emph{NeuroPOL} framework as a surrogate for reliability analysis, we validate its performance on five canonical partial differential equations (PDEs) that are widely used in scientific computing to model a variety of physical phenomena. These PDEs are carefully selected to represent distinct classes of physical processes and, therefore, to comprehensively evaluate the capabilities of NeuroPOL. 

The first example, \textbf{E-I}, involves the \textit{one-dimensional, time-dependent Burgers equation}, which is frequently used as a benchmark for nonlinear diffusion and advection-dominated transport problems. The second example, \textbf{E-II}, focuses on the \textit{one-dimensional, time-dependent Nagumo equation}, which describes reaction-diffusion phenomena and is often used to model excitable media in physics and biology. The third example, \textbf{E-III}, considers the \textit{two-dimensional, time-independent Poisson equation}, which represents an elliptic field problem and forms the foundation for numerous applications involving potential fields, steady-state diffusion, and electrostatics. The fourth example, \textbf{E-IV}, considers a \textit{two-dimensional, time-independent Darcy equation}, which describes laminar flow of a fluid through a porous medium, relating volumetric flux to the pressure gradient via the medium’s permeability. It is widely used to model flow through aquifers, petroleum reservoirs, and other porous systems. Finally, the fifth example, \textbf{E-V}, deals with a \textit{two-dimensional incompressible
Navier-Stokes equation coupled with the energy equation}, which governs buoyancy-driven natural convection under the Boussinesq approximation. It serves as a canonical model for heat transfer in enclosed cavities and other thermo-fluid systems. 

These five examples are deliberately chosen to stress-test NeuroPOL under diverse modeling requirements:  
(1) Burgers’ equation presents nonlinear interactions between convection and diffusion, requiring the surrogate to accurately capture sharp gradients and evolving shock structures;  
(2) the Nagumo equation introduces reaction-driven dynamics, which test the ability of NeuroPOL to resolve traveling wavefronts while maintaining numerical stability;  
(3) the Poisson equation exemplifies a steady-state elliptic problem where spatially smooth solutions must be learned, thereby probing the framework’s ability to generalize from sparse boundary conditions to interior field predictions; (4) the Darcy equation challenges NeuroPOL to map permeability fields to the resulting pressure distributions, testing its ability to generalize across spatially varying material properties; and (5) the Navier–Stokes–energy system stresses the framework’s capacity to resolve tightly coupled velocity–pressure–temperature fields, capturing buoyancy-driven convection, boundary-layer structures, and parameter-dependent flow regimes. Apart from these, the discussed examples also present varied challenges like example E-I and E-II challenge NeuroPOL to learn time-dependent PDEs, whereas E-III to E-V test NeuroPOL's performance on two-dimensional domains. E-IV particularly involves tackling inputs with high intrinsic dimensionality and E-V challenges NeuroPOL to learn coupled PDEs.
By considering these representative PDEs, we ensure that NeuroPOL is evaluated across a broad spectrum of physical processes, covering both time-dependent and time-independent regimes.  

For reliability analysis, different tasks are performed depending on the nature of the PDE. In {E-I} and {E-II}, where the systems evolve over time, we compute the Probability Density Function (PDF) of \textit{First-Passage Failure Time (FPFT)}, which quantifies the probability distribution of the time at which a system trajectory first crosses a specified failure threshold. In contrast, in {E-III}, E-IV and E-V, the focus is on the direct computation of the \textit{failure probability}, defined as the likelihood of the solution exceeding a critical limit within the domain of interest. Each case is solved using the \textbf{NeuroPOL} architecture, and the results are benchmarked against two reference approaches: the \textit{vanilla Physics-Informed Wavelet Neural Operator (PIWNO)}, which provides a neural operator baseline without the enhancements introduced in NeuroPOL, and \textit{Monte Carlo simulations}, which are treated as the gold standard for reliability quantification. Comparisons are drawn in two dimensions: first, the \textit{accuracy of predictive solutions} obtained by NeuroPOL relative to the ground truth, and second, the \textit{fidelity of reliability estimates} derived from these predictions when compared with the results obtained using Monte Carlo methods.  

The \textbf{NeuroPOL} architecture employed in these examples is configured with a variable number of iterative layers, typically ranging between three and five, depending on the complexity of the governing PDE. After each iterative layer, the corresponding \textit{Variable Spiking Neuron (VSN)} activation functions are applied, except for the final iterative layer and the two dense projection layers that map between the functional and operator spaces. Optimization of the network parameters is carried out using the {ADAM algorithm} with an \textbf{initial learning rate} of $10^{-3}$, which is \textit{geometrically decayed} by a factor of $0.75$ every $50$ epochs to ensure convergence stability. Furthermore, a \textit{weight decay} of $10^{-6}$ is introduced to regularize the model and mitigate overfitting. 
The performance of NeuroPOL across all five examples is evaluated using the \textit{Normalized Mean Squared Error (NMSE)}, computed on withheld test samples. This metric provides a consistent and rigorous basis for quantifying predictive accuracy while enabling fair comparisons across PDEs of varying dimensionality and complexity.
\subsection{Example E-I: Nonlinear Diffusion in Reactive Flows}
As a model for nonlinear convective-diffusive transport, the one-dimensional viscous Burgers’ equation appears frequently in simulations of shock wave formation, fluid turbulence, and traffic dynamics. The governing PDE is defined as,
\begin{equation}
\frac{\partial u(x, t)}{\partial t} + \frac{1}{2} \frac{\partial u^2(x, t)}{\partial x} = \nu \frac{\partial^2 u(x, t)}{\partial x^2}, \quad x \in [0, 1],\ t \in [0, 1],
\end{equation}
with the viscosity coefficient being fixed at $\nu = 0.1$. The PDE is subjected to periodic boundaries and an initial condition,
\begin{equation}
u(x,0) = \cos(\zeta \pi x) + \sin(\eta \pi x), \quad \zeta, \eta \sim \mathbb{U}(0.5, 1.5).
\end{equation}
The model's objective is to learn the operator mapping initial condition to the solution on the whole grid, i.e., $u(x,0) \mapsto u(x, t)$. The inputs in this example are parameterized by two independent variables $\zeta$ and $\eta$, giving the input space an intrinsic dimensionality of two. Ground truth solutions are computed using a MATLAB PDE solver across a grid of $81 \times 81$ in space and time. NeuroPOL and PIWNO models are trained for 300 epochs. 300 samples of initial conditions are used to train the deep learning models, and 1500 samples are used for testing the performance of various models. We note that no labelled dataset is required and hence, the NeuroPOL is trained in a simulation free setup.
\begin{figure}[ht!]
    \centering
    \includegraphics[width=1\linewidth]{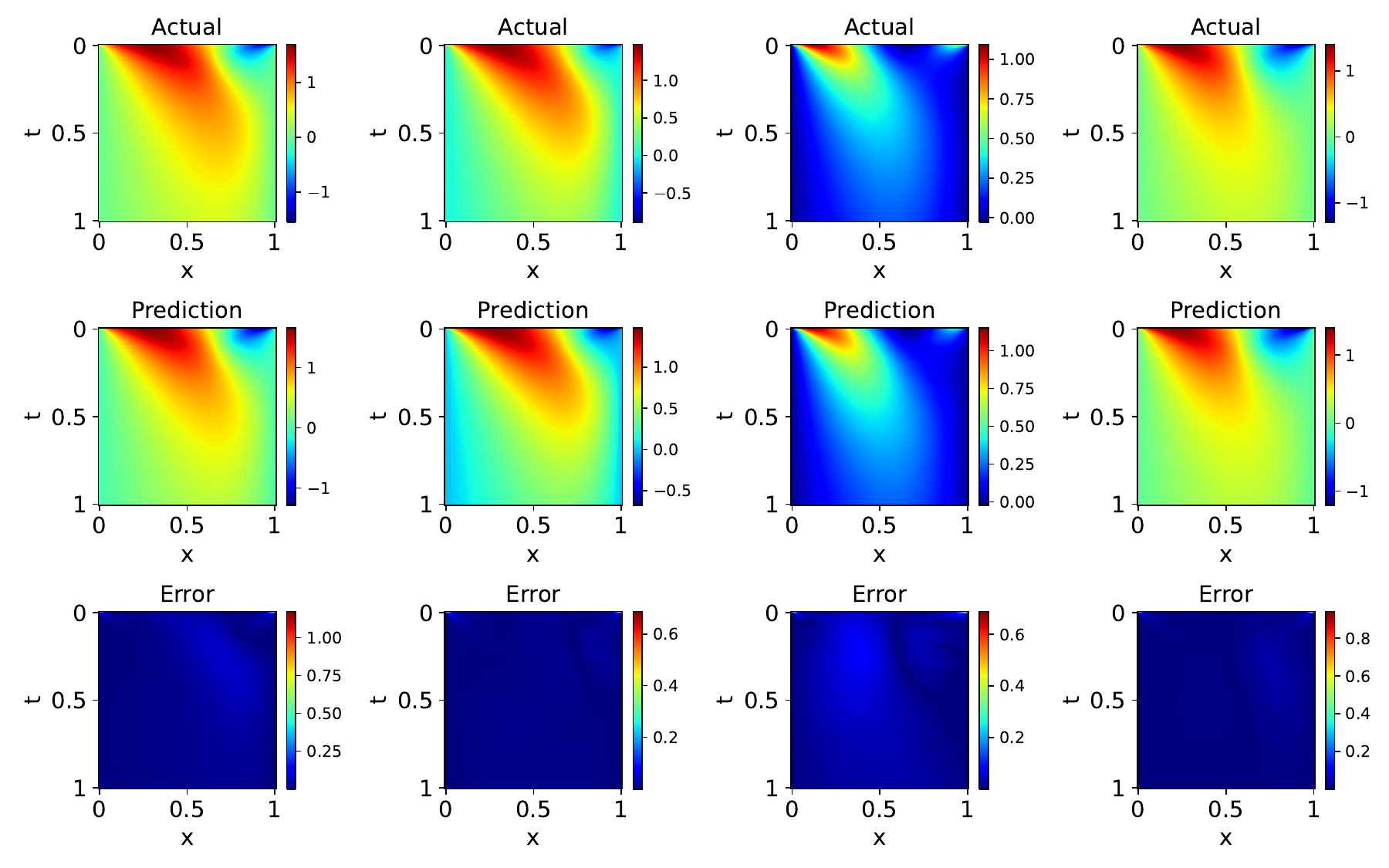}
    \caption{Representative examples from the test dataset comparing NeuroPOL predictions with ground truth for Burgers PDE in example, E-I. The first row shows ground truth data, the second row the corresponding predictions, and the third row the absolute error.}
    \label{fig: burgers pred vspiwno}
\end{figure}

Fig. \ref{fig: burgers pred vspiwno} compares the NeuroPOL predictions against the ground truth (actual results). As can be observed, the predictions closely follow the ground truth. An NMSE of 0.31 \% was observed with a spiking activity of approximately 18\% and 5\% in two the two VSN layers, used within the NeuroPOL architecture. Despite sparse communication, the NMSE observed is well within a reasonable range and compares well against the observed NMSE of 0.13\% observed when using vanilla PIWNO. 
A variant of NeuroPOL was also tested wherein an extra VSN layer was used after the uplifting operator and before the first iterative layer $L_1$. An NMSE of 0.13\% was observed for this variant, which is similar to that observed when using vanilla PIWNO.
\begin{figure}[ht!]
    \centering

    \begin{subfigure}{0.49\textwidth}
        \centering
        \includegraphics[width=\textwidth]{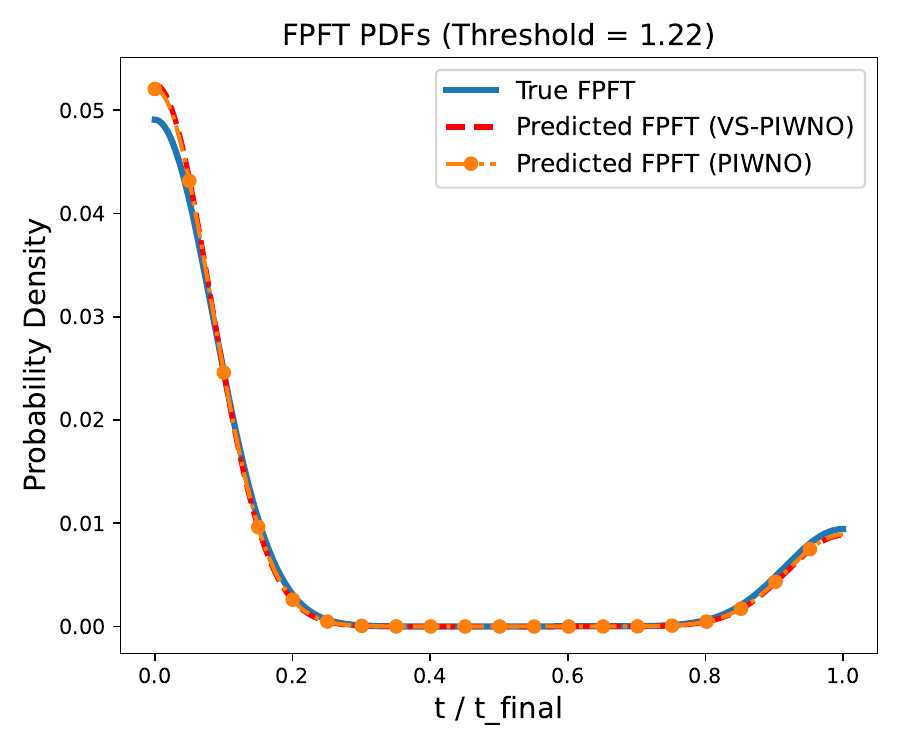}
        \caption{Threshold = 1.22.}
    \end{subfigure}
    \hfill
    \begin{subfigure}{0.49\textwidth}
        \centering
        \includegraphics[width=\textwidth]{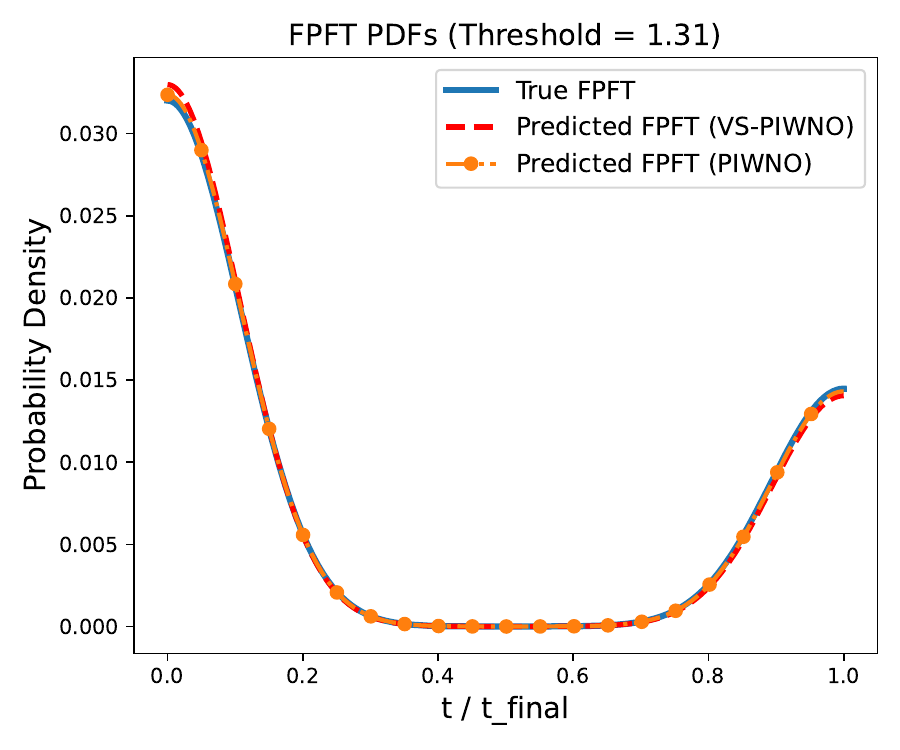}
        \caption{Threshold = 1.31.}
    \end{subfigure}
    \hfill
    \begin{subfigure}{0.49\textwidth}
        \centering
        \includegraphics[width=\textwidth]{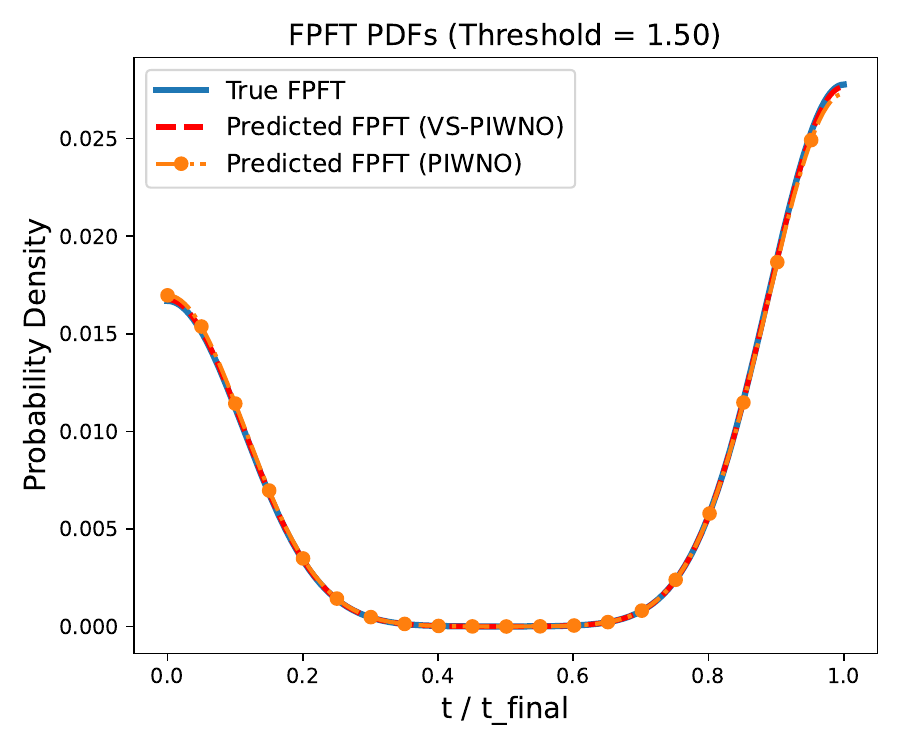}
        \caption{Threshold = 1.50.}
    \end{subfigure}

    \caption{Comparison of FPFT PDFs obtained corresponding to different thresholds for the Burgers PDE in example, E–I. Each sub-figure contrasts the true PDF with those obtained using NeuroPOL predictions and PIWNO predictions.}
    \label{fig: burgers fpft_comparison}
\end{figure}
Fig.~\ref{fig: burgers fpft_comparison} presents the probability density functions (PDFs) of the FPFT computed for the current example across different threshold values. As can be observed, the PDFs obtained using NeuroPOL predictions closely align with the true PDFs estimated via MCS as well as those derived from the vanilla PIWNO framework. This demonstrates the capability of the proposed NeuroPOL framework to deliver highly accurate results while simultaneously reducing energy consumption, as evidenced by the spiking activity observed in the VSN layers.

\subsection{Example E-II: Neural Impulse Dynamics in Axons}
As the second example, we tackle the Nagumo equation, a classical model for electrical signal propagation along neuronal fibers. It combines nonlinear reaction kinetics with diffusion and is defined as,
\begin{equation}
\frac{\partial u}{\partial t} - \epsilon \frac{\partial^2 u}{\partial x^2} = u(1 - u)(u - \alpha), \quad x \in [0, 1],\ t \in [0, 1].
\end{equation}
The reaction parameter is set to $\alpha = -0.5$, and diffusion is governed by $\epsilon = 1$. The PDE is subjected to zero-flux boundary conditions, 
\begin{equation}
\left. \frac{\partial u}{\partial x} \right|_{x=0} = \left. \frac{\partial u}{\partial x} \right|_{x=1} = 0,
\end{equation}
and randomized initial conditions $u(x, 0) = u_0(x)$, where $u_0(x)$ is sampled from a zero-mean Gaussian random field with covariance,
\begin{equation}
\mathcal{K}(x, x') = \sigma^2 \exp\left(-\frac{(x - x')^2}{2 l^2} \right), \quad \sigma = 0.1,\ l = 0.1.
\end{equation}
Although the Gaussian random field, used to generate the input in this example, is theoretically infinite-dimensional, eigenvalue analysis shows that the first ten leading eigenvalues capture 99\% of the energy, giving the input space an intrinsic dimensionality of ten.
Reference solutions are generated using a semi-implicit Euler scheme combined with a central difference method. The model is trained on a $65 \times 65$ grid for 400 epochs. 800 input samples are used for training, and 1200 samples are used for testing the various deep learning models. Similar to the previous example, the model is trained in a simulation free setup

\begin{figure}[ht!]
    \centering
    \includegraphics[width=1\linewidth]{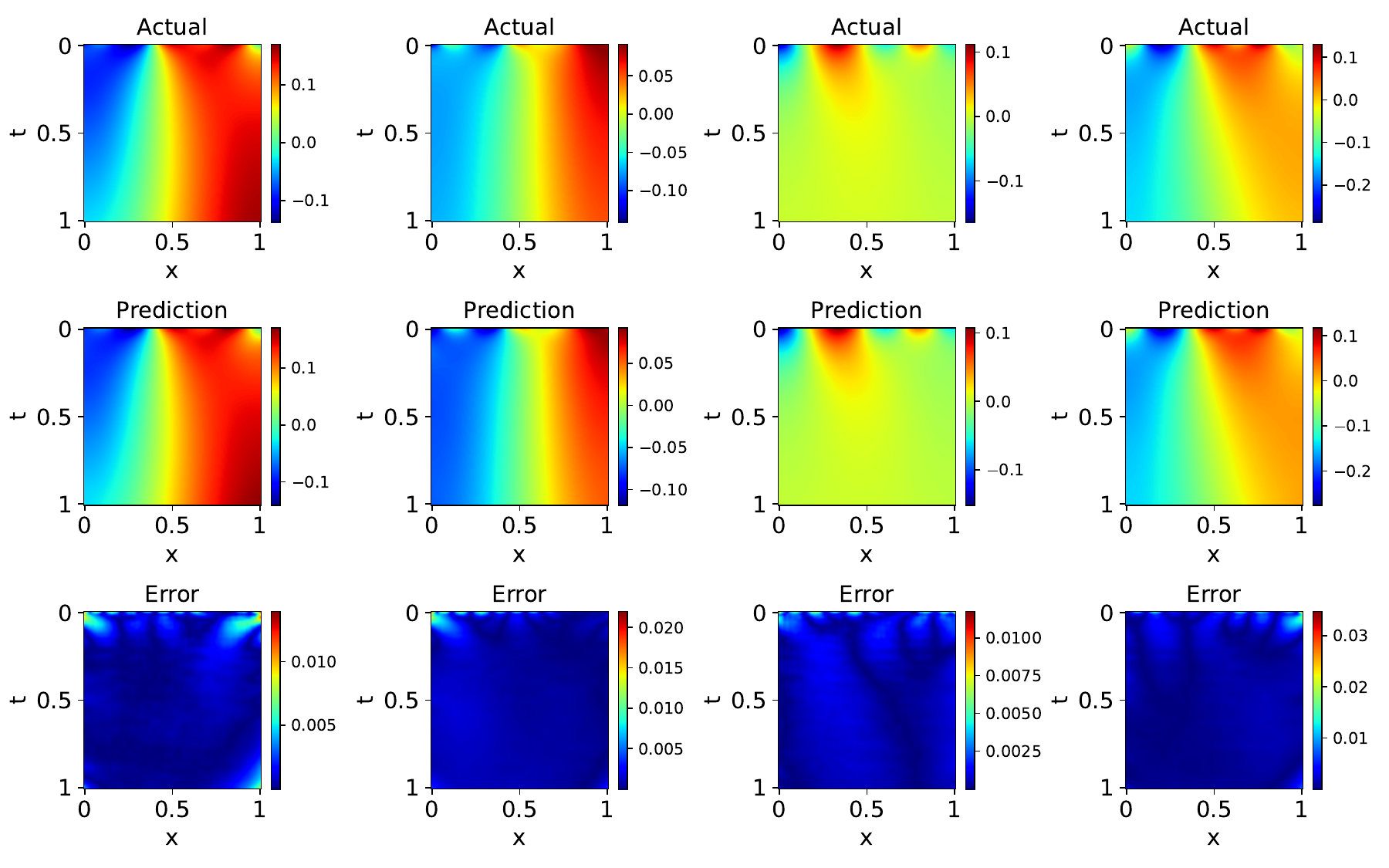}
    \caption{Representative examples from the test dataset comparing NeuroPOL predictions with ground truth for Nagumo PDE in example, E-II. The first row shows ground truth data, the second row the corresponding predictions, and the third row the absolute error.}
    \label{fig: nagumo vspiwno}
\end{figure}
Fig. \ref{fig: nagumo vspiwno} compares the NeuroPOL predictions against the ground truth. Similar to the previous example, the predictions closely follow the ground truth. An NMSE of 0.07\% is observed when using NeuroPOL against an NMSE value of 0.05\% when using vanilla PIWNO. Spiking activity observed when using NeuroPOL is approximately 73\%, 48\%, 75\%, and 68\% across the four VSN layers used in NeuroPOL.  Similar to the previous example, a case where an extra VSN layer is used outside the uplifting operator, an NMSE value of 0.07\% was observed.

\begin{figure}[ht!]
    \centering

    \begin{subfigure}{0.49\textwidth}
        \centering
        \includegraphics[width=\textwidth]{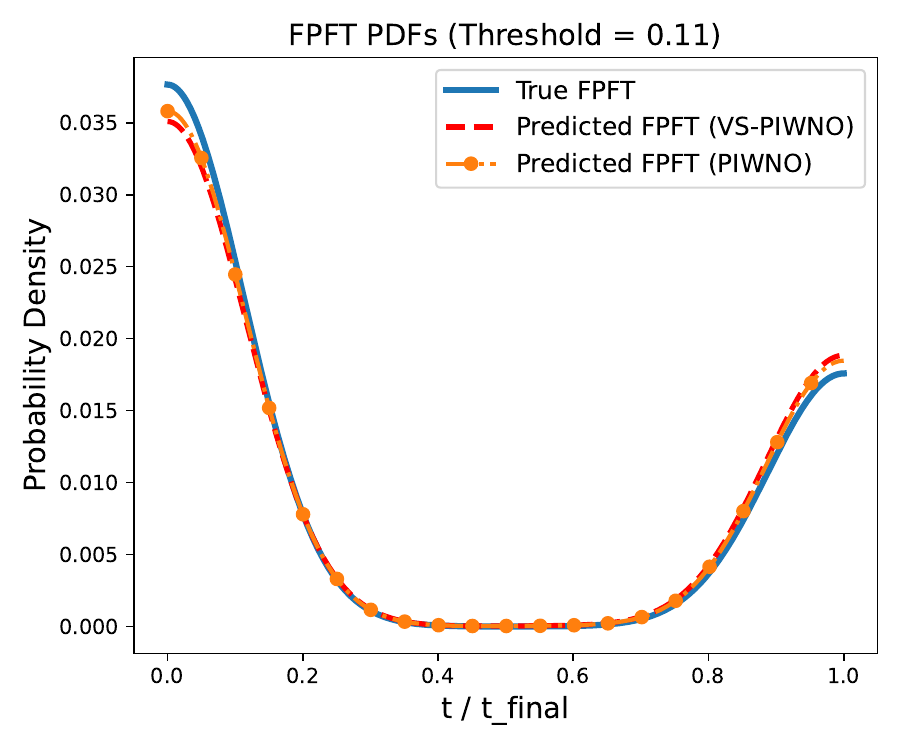}
        \caption{Threshold = 0.11.}
    \end{subfigure}
    \hfill
    \begin{subfigure}{0.49\textwidth}
        \centering
        \includegraphics[width=\textwidth]{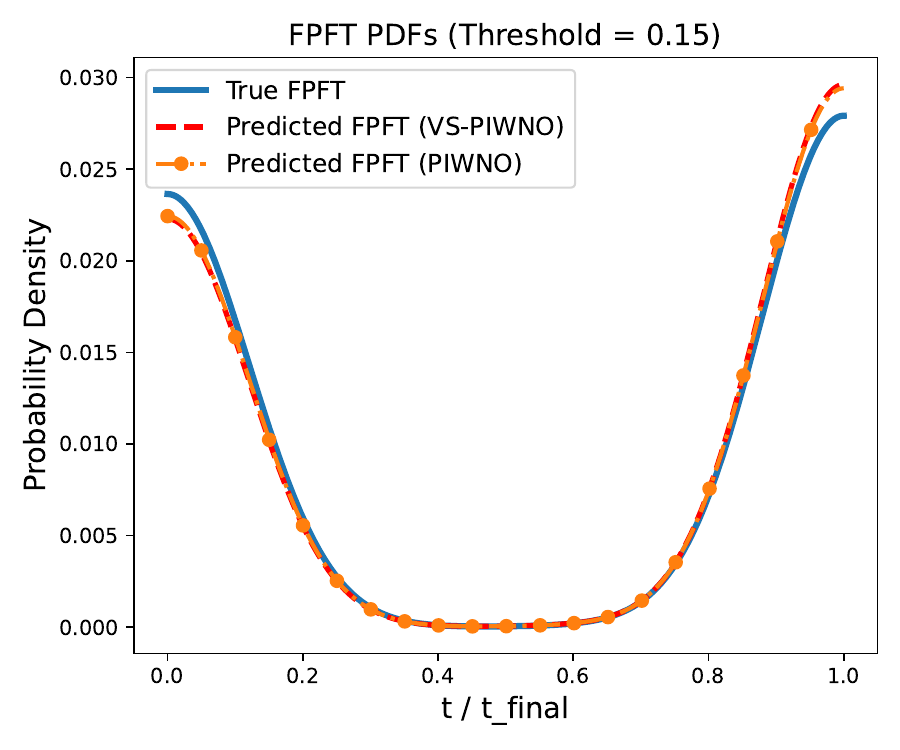}
        \caption{Threshold = 0.15.}
    \end{subfigure}
    \hfill
    \begin{subfigure}{0.49\textwidth}
        \centering
        \includegraphics[width=\textwidth]{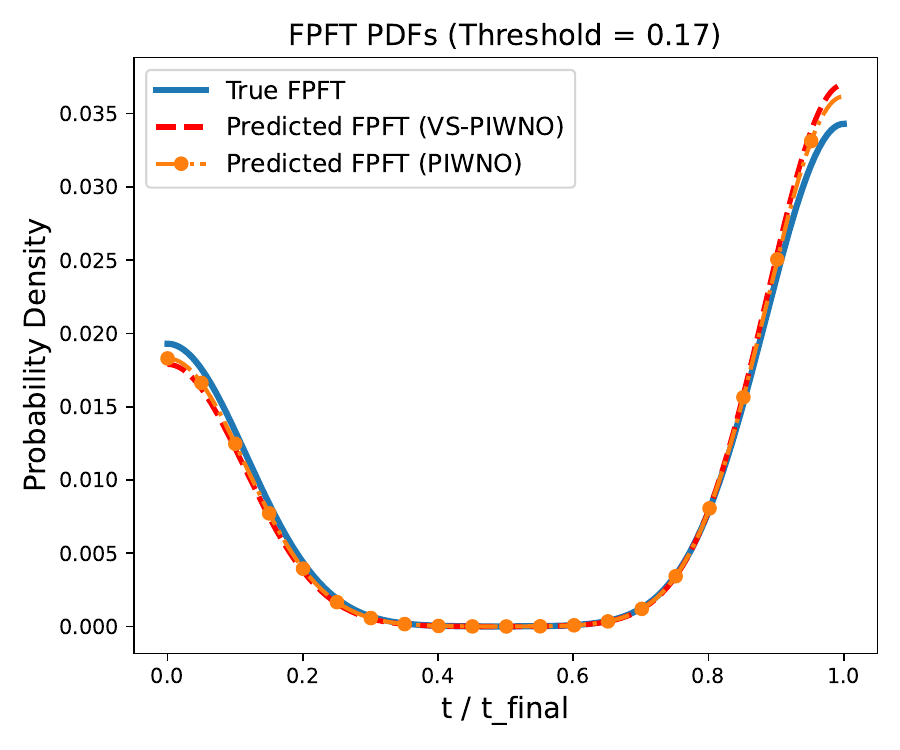}
        \caption{Threshold = 0.17.}
    \end{subfigure}

    \caption{Comparison of FPFT PDFs obtained corresponding to different thresholds for the Nagumo PDE in example, E–II. Each sub-figure contrasts the true PDF with those obtained using NeuroPOL predictions and PIWNO predictions.}
    \label{fig: nagumo fpft_comparison}
\end{figure}
Fig. \ref{fig: nagumo fpft_comparison} shows the PDFs for FPFT obtained for different threshold values. We observe that the PDFs obtained using NeuroPOL follow the ground truth obtained using MCSs closely and compare well against the PDFs obtained using vanilla PIWNO, despite sparse communication. This again is indicative of the capability of the proposed NeuroPOL is yielding an accurate and energy efficient alternative to classical neural operators.
\subsection{Example E-III: Steady-State Field Estimation in Electrodynamics}
The third example focuses on a two-dimensional, time-independent Poisson equation, frequently used in electrostatics and steady-state heat conduction modeling. The PDE is defined as,
\begin{equation}
\frac{\partial^2 u(x, y)}{\partial x^2} + \frac{\partial^2 u(x, y)}{\partial y^2} = f(x, y), \quad (x, y) \in [-1, 1]^2,
\end{equation}
The PDE is subjected to homogeneous Dirichlet conditions,
\begin{equation}
u(x, y) = 0, \quad \text{on} \ \partial \Omega.
\end{equation}
For a source term of the form,
\begin{multline}
f(x, y) = 16 \beta \pi^2 \left( \cos(4\pi y)\sin(4\pi x) + \sin(4\pi x)(\cos(4\pi y) - 1) \right) \\- \alpha \pi^2 \left( \cos(\pi y)\sin(\pi x) + \sin(\pi x)(\cos(\pi y) + 1) \right),
\end{multline}
the exact solution is given as,
\begin{equation}
u(x, y) = \alpha \sin(\pi x)(1 + \cos(\pi y)) + \beta \sin(2\pi x)(1 - \cos(2\pi y)),
\end{equation}
Coefficients $\alpha$ and $\beta$ are sampled from a uniform distribution $\mathbb U(-2, 2)$. The operator learning task is to map the source field to the solution field, i.e., $f(x, y) \mapsto u(x, y)$. Similar to the first example, the inputs in this example are parameterized by two independent variables $\alpha$ and $\beta$, giving the input space an intrinsic dimensionality of two. The model is trained on $65 \times 65$ spatial grids for 350 epochs. 500 samples are used for training, and 1500 samples are used for testing. 

\begin{figure}[ht!]
    \centering
    \includegraphics[width=1\linewidth]{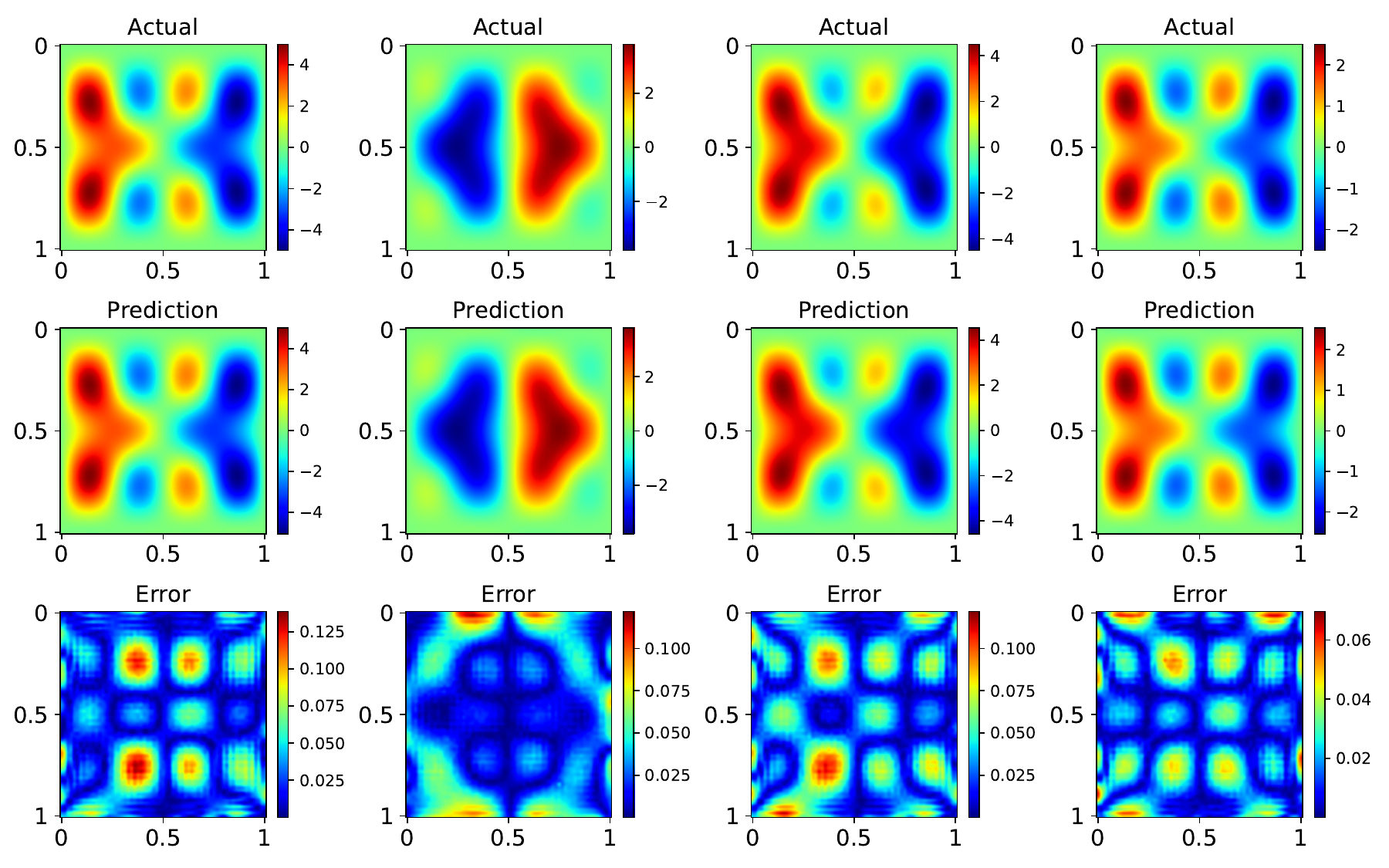}
    \caption{Representative examples from the test dataset comparing NeuroPOL predictions with ground truth for Poisson PDE in example, E-III. The first row shows ground truth data, the second row the corresponding predictions, and the third row the absolute error.}
    \label{fig: poissons NeuroPOL predictions}
\end{figure}
Fig. \ref{fig: poissons NeuroPOL predictions} shows the NeuroPOL predictions compared against the ground truth, and as can be seen, the predictions closely follow the ground truth. An NMSE of 0.049\% was observed when using NeuroPOL against an NMSE of 0.042\% when using vanilla PIWNO. A spiking activity of approximately 60\%, 50\%, 55\%, 65\%, and 60\% was observed in the five VSN layers used in the NeuroPOL model. An NMSE of 0.050\% was observed when using an extra VSN layer around the uplifting operator.

\begin{table}[ht!]
\centering
\caption{Comparison of estimated failure probabilities, $p_f$, corresponding to different threshold values for the Poisson PDE in example, E–III. The table contrasts ground truth values with those obtained using NeuroPOL predictions and PIWNO predictions.}
\begin{tabular}{cccc}
\hline
{Threshold} & {True $p_f$} & {NeuroPOL $p_f$} & {PIWNO $p_f$} \\
\hline
1.66 & 0.870 & 0.875 & 0.874 \\
1.94 & 0.814 & 0.822 & 0.822 \\
2.21 & 0.760 & 0.765 & 0.767 \\
\hline
\end{tabular}
\label{tab: pfailure poissons}
\end{table}
Table \ref{tab: pfailure poissons} shows the probability of failure obtained corresponding to different threshold values. As can be seen, the values obtained using NeuroPOL closely follow the ground truth obtained using MCSs.
\subsection{Example E-IV: Flow Through Porous Media}
In the fourth example, we deal with the problem of flow through porous media. This phenomenon is often encountered in the fields of fluid mechanics, soil engineering,  reservoir engineering, and the petroleum industry. Darcy’s equation is used to model the said phenomenon and the same is defined as  
\begin{equation}
\begin{gathered}
-\nabla \cdot \big( a(x,y)\,\nabla u(x,y) \big) = 1, 
\quad (x,y) \in (0,1)^2,\\ 
\quad u(x,y) = 0, \quad (x,y) \in \partial(0,1)^2,
\end{gathered}
\end{equation}
where $a(x,y)$ denotes the permeability field and $u(x,y)$ the resulting pressure field. The permeability field $a(x,y)$ is modeled as a Gaussian random field as, $
a \sim \mathcal{N}\!\left(0,\,2.0245(-\Delta + 2.56I)^{-1.6}\right)
$, where $\Delta$ is the Laplacian operator with zero Neumann boundary conditions. A spatial resolution of $64 \times 64$ is considered, and the inputs have an effective dimension of approximately 305 corresponding to a variance threshold of 99\%. In this case, the objective is to learn the operator mapping the permeability field to the final solution, i.e., $a(x,y) \mapsto u(x,y)$. 1000 samples are used for training, and 1000 samples are used for testing and reliability analysis.  

\begin{figure}[ht!]
    \centering
    \includegraphics[width=1\linewidth]{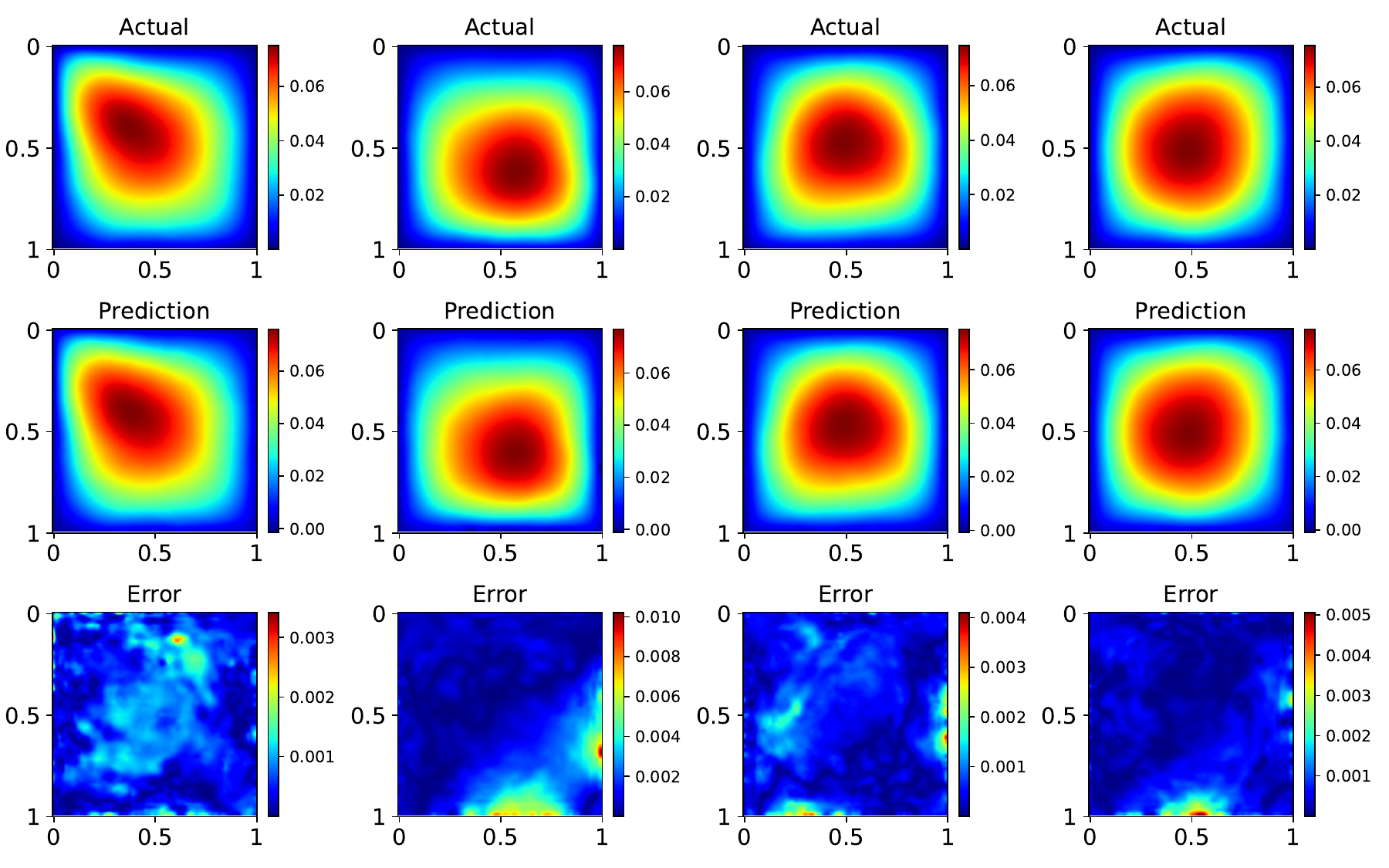}
    \caption{Representative examples from the test dataset comparing NeuroPOL predictions with ground truth for Darcy PDE in example, E-IV. The first row shows ground truth data, the second row the corresponding predictions, and the third row the absolute error.}
    \label{fig: darcy vspiwno}
\end{figure}
An NMSE of 0.08\% is observed when using NeuroPOL against an NMSE value of 0.06\% when using vanilla PIWNO. The predictions give a good approximation of the ground truth and the same is confired by Fig. \ref{fig: darcy vspiwno}.
A spiking activity in the range of 60\% to 90\% was observed in the three spiking layers of the NeuroPOL network. 
\begin{table}[ht!]
\centering
\caption{Comparison of estimated failure probabilities, $p_f$, corresponding to different threshold values for the Darcy PDE in example, E–IV. The table contrasts ground truth values with those obtained using NeuroPOL predictions and PIWNO predictions.}
\begin{tabular}{cccc}
\hline 
{Threshold} & {True $p_f$} & {NeuroPOL $p_f$} & {PIWNO $p_f$} \\
\hline 
0.076 & 0.297 & 0.283 & 0.362 \\
0.078 & 0.122 & 0.128 & 0.164 \\
0.080 & 0.035 & 0.036 & 0.044 \\
\hline
\end{tabular}
\label{tab: pfailure darcy}
\end{table}
Table \ref{tab: pfailure darcy} shows the probability of failure obtained corresponding to different threshold values. The values obtained using NeuroPOL closely follow the ground truth obtained using MCSs, and in this example, NeuroPOL performs better than the vanilla PIWNO network in estimating the probability of failure.
\subsection{Example E-V: Natural Convection Based Cooling in Enclosed Cavities}
Heat transfer through natural convection inside enclosed cavities represents a key process in thermo-fluid dynamics. The mechanism, driven by density variations associated with temperature differences, directly impacts how thermal energy is distributed in confined spaces. Because of this, it plays a critical role in engineering applications, including electronic device cooling and building climate control. The fluid motion is induced by buoyancy forces that originate from temperature-dependent density variations, leading to coupled velocity and temperature fields. For steady-state conditions with moderate temperature gradients, this behavior is effectively captured by the Boussinesq approximation, where the system is represented through the two-dimensional incompressible Navier--Stokes equations coupled with an energy equation. For a square cavity of side length $H$, the governing equations are expressed as
\begin{equation}
\begin{gathered}
\frac{\partial U}{\partial X} + \frac{\partial V}{\partial Y} = 0,\\
U \frac{\partial U}{\partial X} + V \frac{\partial U}{\partial Y} 
= -\frac{\partial P}{\partial X} 
+ \left(\frac{\mathrm{Pr}}{\mathrm{Ra}}\right)^{1/2} 
\left(\frac{\partial^2 U}{\partial X^2} + \frac{\partial^2 U}{\partial Y^2}\right),\\
U \frac{\partial V}{\partial X} + V \frac{\partial V}{\partial Y} 
= -\frac{\partial P}{\partial Y} 
+ \left(\frac{\mathrm{Pr}}{\mathrm{Ra}}\right)^{1/2} 
\left(\frac{\partial^2 V}{\partial X^2} + \frac{\partial^2 V}{\partial Y^2}\right) + \theta,\\
U \frac{\partial \theta}{\partial X} + V \frac{\partial \theta}{\partial Y} 
= \left(\frac{1}{\mathrm{Ra}\,\mathrm{Pr}}\right)^{1/2}
\left(\frac{\partial^2 \theta}{\partial X^2} + \frac{\partial^2 \theta}{\partial Y^2}\right),
\end{gathered}
\end{equation}
where, $U$ and $V$ denote the dimensionless velocity components, $P$ the scaled pressure, and $\theta$ the non-dimensional temperature. $\mathrm{Ra} = 3000$ and $\mathrm{Pr} = 0.71$ are the Rayleigh an Prandtl numbers respectively. The spatial coordinates $X$ and $Y$, along with the velocity components, pressure, and temperature are non-dimensionalized as,
\begin{equation}
\begin{gathered}
(X, Y) = \left(\frac{x}{H}, \frac{y}{H}\right), \quad
(U, V) = \left(\frac{u}{\sqrt{\beta g (T_h - T_c) H}}, \frac{v}{\sqrt{\beta g (T_h - T_c) H}}\right),\\
\theta = \frac{T - T_c}{T_h - T_c}, \quad 
P = \frac{p}{\rho g \beta (T_h - T_c)H},
\end{gathered}
\end{equation}
where $g$ is the gravitational acceleration, $\beta$ is the thermal expansion coefficient.
%
%
%
The boundary conditions for the cavity are imposed as no-slip on all walls i.e. $U = V = 0$. On the left boundary, $X=0$, a spatially varying temperature profile $\theta = f'(Y)$ is applied. The right wall, $X=1$, is maintained at $\theta=0$, while adiabatic conditions are prescribed at the top and bottom boundaries, i.e., $\partial \theta / \partial Y = 0$. The uncertainty in the system is modeled through the left-wall temperature profile, described as,
\begin{equation}
\theta(Y) = f'(Y) = n_1 Y (1-Y)\left[n_2 \cos(\pi Y) + n_3 \sin(\pi Y)\right]^2,
\end{equation}
where $n_1$, $n_2$, and $n_3$ are the random parameters characterizing the stochastic input. As a result, the inputs in this example have an effective dimension of three. We will look at the probability of failure in the event that the controlling temperature crosses a certain threshold, as exceeding thresholds in this case may result in degradation, reduced efficiency, or failure of materials and components.
The computational domain is discretized on a $101 \times 101$ grid, and 280 samples are used for training, while 1100 samples are used for prediction and testing the performance in gauging the reliability of the system. In the current example, true labels are also used while training along with the governing physics.

\begin{figure}[ht!]
    \centering
    \includegraphics[width=1\linewidth]{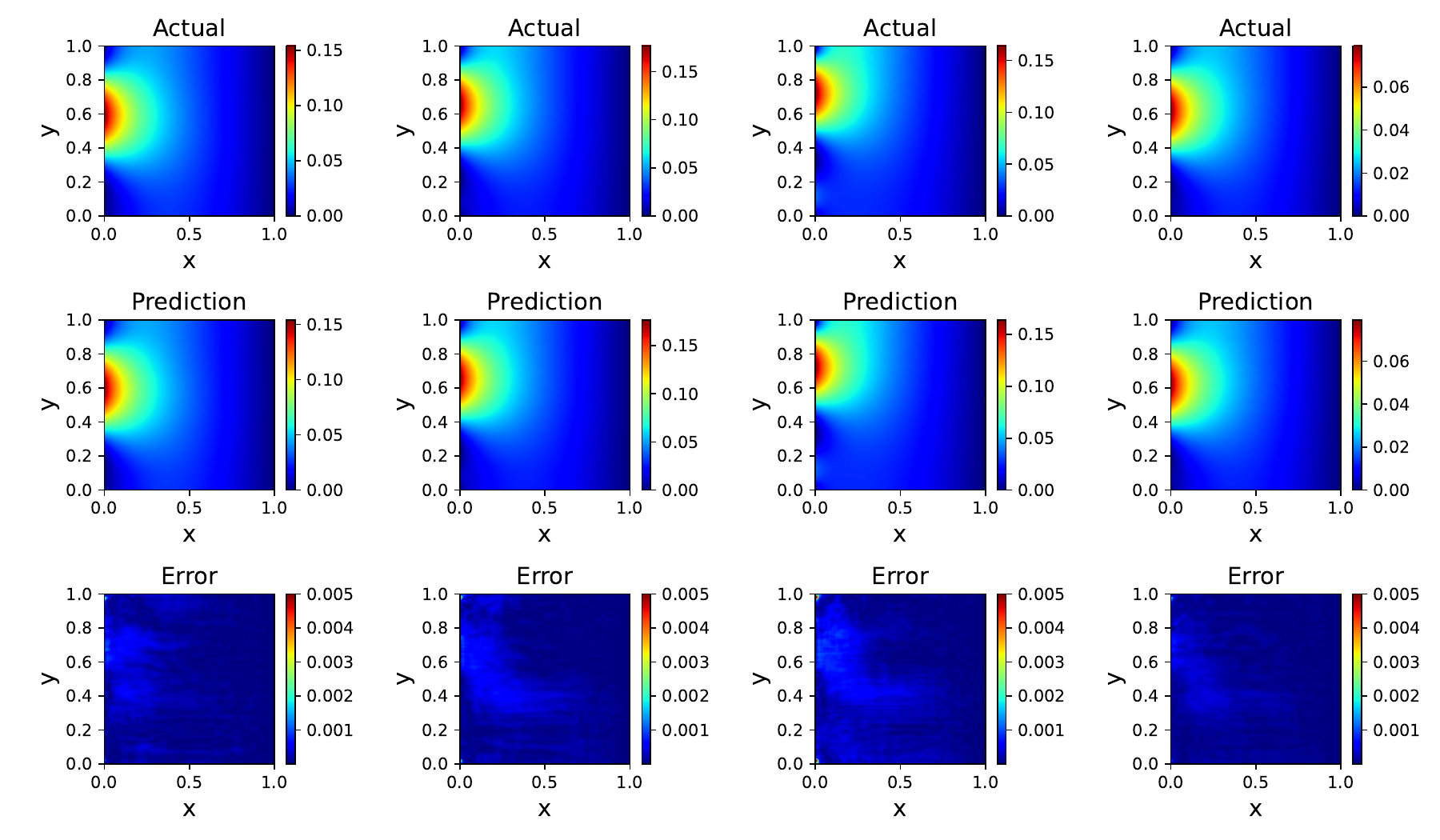}
    \caption{Representative examples from the test dataset comparing NeuroPOL predictions with ground truth for coupled Navier-Stokes and energy PDE system in example, E-V. The first row shows ground truth data, the second row the corresponding predictions, and the third row the absolute error.}
    \label{fig: natural vspiwno}
\end{figure}
Fig. \ref{fig: natural vspiwno} compares the NeuroPOL predictions against the ground truth. Similar to the previous example, the predictions closely follow the ground truth. An NMSE of 0.0065\% is observed when using NeuroPOL against an NMSE value of 0.0034\% when using vanilla PIWNO. 
A spiking activity in the range of 65\% to 90\% was observed in various spiking layers of the NeuroPOL network. 
\begin{table}[ht!]
\centering
\caption{Comparison of estimated failure probabilities, $p_f$, corresponding to different threshold values for the coupled Navier-Stokes and energy PDE system in example, E–V. The table contrasts ground truth values with those obtained using NeuroPOL predictions and PIWNO predictions.}
\begin{tabular}{cccc}
\hline 
{Threshold} & {True $p_f$} & {NeuroPOL $p_f$} & {PIWNO $p_f$} \\
\hline 
0.069 & 0.372 & 0.372 & 0.373 \\
0.087 & 0.283 & 0.283 & 0.284 \\
0.139 & 0.127 & 0.125 & 0.127 \\
\hline
\end{tabular}
\label{tab: pfailure natural}
\end{table}
Table \ref{tab: pfailure natural} shows the probability of failure obtained corresponding to different threshold values. As seen in previous examples, despite sparse communication, the values obtained using NeuroPOL closely follow the ground truth obtained using MCSs.
\section{Conclusion}\label{section: conclusion}
Reliability analysis of engineering systems under uncertainty often requires repeated evaluations of complex physical models governed by high-dimensional, nonlinear, and multiphysics PDEs. While Monte Carlo simulations remain the gold standard for such analyses, their prohibitive computational cost severely limits their practicality, particularly in scenarios demanding real-time decision-making or deployment on energy-constrained platforms. Surrogate models therefore become essential; however, conventional approaches struggle to simultaneously achieve high predictive accuracy, physics consistency, scalability, and energy efficiency. In this work, we introduced NeuroPOL, the first-ever neuroscience-inspired physics-informed operator learning framework for reliability analysis. NeuroPOL seamlessly integrates Variable Spiking Neurons (VSNs) within a physics-informed operator learning backbone, combining the rigor of physics-based modeling with the efficiency of sparse, event-driven communication. By leveraging a purely physics-grounded loss, NeuroPOL eliminates the need for labeled data while ensuring that the learned operators remain consistent with the governing PDEs. The incorporation of VSNs into a Wavelet Neural Operator architecture enables selective replacement of continuous activations with spiking dynamics, striking an optimal balance between predictive fidelity and energy efficiency.

Extensive experiments on five canonical benchmarks, the Burgers equation, the Nagumo equation, a two-dimensional Poisson equation, a two-dimensional Darcy equation, and a two-dimensional incompressible
Navier-Stokes equation coupled with the energy equation demonstrates that NeuroPOL delivers highly accurate state predictions, achieving normalized mean squared errors as low as 0.0065\%, while maintaining average spiking rates significantly below full activation. More importantly, NeuroPOL accurately predicts failure probabilities and First Passage Failure Time (FPFT) distributions, closely matching results obtained via Monte Carlo simulations and conventional physics-informed operators, thereby validating its trustworthiness for reliability analysis. Furthermore, layerwise spiking activity as low as 5\% indicates a substantial reduction in synaptic operations, underscoring NeuroPOL’s potential for energy-efficient deployment on neuromorphic and edge-computing platforms.
Overall, NeuroPOL establishes a new paradigm for energy-aware, physics-consistent surrogate modeling tailored for reliability analysis. Its unique combination of predictive accuracy and energy efficiency makes it a promising candidate for real-time reliability assessment and seamless integration within digital twin ecosystems. Future work will focus on extending NeuroPOL to high-dimensional, irregular, and multiphysics domains, as well as conducting hardware-in-the-loop experiments to quantify actual energy savings and latency benefits on neuromorphic platforms.

\section*{Acknowledgment}
SG acknowledges the financial support received from the Ministry of Education, India, in the form of the Prime Minister's Research Fellows (PMRF) scholarship. SC acknowledges the financial support received from the Anusandhan National Research Foundation (ANRF) via grant no. CRG/2023/007667 and from the Ministry of Port and Shipping via letter no. ST-14011/74/MT (356529).

\end{document}